\ifdefined\pdfminorversion\pdfminorversion=7\fi
\documentclass[runningheads]{llncs}

\usepackage[T1]{fontenc}
\usepackage{graphicx}
\usepackage{amsmath,amssymb}
\usepackage{booktabs}
\usepackage{multirow}
\usepackage{array}
\usepackage{algorithm}
\usepackage{algpseudocode}
\usepackage{placeins}
\usepackage{float}
\usepackage{url}
\usepackage[hidelinks]{hyperref}
\hypersetup{
  pdftitle={Genetic Programming with Behaviour-based Niching for Learning Guided Local Search in Vehicle Routing Problems},
  pdfauthor={Saining Liu, Yi Mei, Mengjie Zhang},
  pdfsubject={Behaviour-based population management for Genetic Programming Guided Local Search},
  pdfkeywords={Vehicle routing, Genetic programming, Guided local search, Behaviour-based niching}
}
\usepackage{etoolbox}
\AtBeginEnvironment{thebibliography}{\setlength{\itemsep}{0pt}\setlength{\parsep}{0pt}\setlength{\parskip}{0pt}}

\begin{document}

\title{Genetic Programming with Behaviour-based Niching for Learning Guided Local Search in Vehicle Routing Problems}
\titlerunning{Learning GPGLS with Behaviour-based Niching}

\author{Saining Liu \and Yi Mei \and Mengjie Zhang}
\authorrunning{S. Liu et al.}
\institute{School of Engineering and Computer Science,\\
Victoria University of Wellington, Wellington, New Zealand}

\maketitle

\begin{abstract}
Genetic Programming Guided Local Search (GPGLS) learns utility functions that guide local search for vehicle routing. Its evolving programs can have similar fitness while inducing different search behaviour, making fitness alone an incomplete basis for population diversity management. We propose GPGLS with Behaviour-based Niching (BN-GPGLS), which characterises programs through six operator-level descriptors collected during local search. A current-generation archive selects fitness-competitive, compact representatives from strata of a behaviour score. Fixed policies use archive parents continuously, whereas adaptive policies activate them using training-fitness and standardised behaviour-dispersion signals, optionally with a tree-size condition. We compare four behaviour-based variants with a no-archive GPGLS control and fitness-based niching over 30 seed-matched runs on generated 200-customer instances. BN-Adaptive achieves the best descriptive average rank on a separate 90-instance monitoring set; aggregate routing-cost differences are small. All five archive policies produce lower final-population median tree sizes than the GPGLS control, with paired Wilcoxon comparisons remaining significant after Holm adjustment. These results identify useful solution-quality and program-size trade-offs within the evaluated setting, without attributing the size reductions to behaviour representation alone.
\end{abstract}

\section{Introduction}

Vehicle Routing Problems (VRPs) are fundamental combinatorial optimisation problems with broad applications in logistics, transportation, distribution planning, and supply chain management~\cite{clarke1964scheduling,dantzig1959truck,laporte2009fifty,toth2014vehicle}. The general goal of VRPs is to construct a set of feasible vehicle routes that serve all required customers while minimising routing cost, typically subject to operational constraints such as vehicle capacity, route feasibility, and service requirements. Since VRPs are NP-hard, exact methods are usually impractical for large instances, and heuristic or metaheuristic approaches remain the dominant solution techniques for obtaining high-quality solutions within practical computational budgets~\cite{toth2014vehicle,vidal2022hgs}. Recent learning-assisted combinatorial optimisation methods further show that routing and local-search procedures can benefit from automatically learned guidance~\cite{bengio2021machine,hudson2021graph,kool2018attention}. Genetic Programming Guided Local Search (GPGLS) is one such hyper-heuristic framework that automatically evolves utility functions for Guided Local Search (GLS), reducing the need for manually designed penalty rules and enabling problem-dependent search guidance~\cite{liu2024gpgls,tsang1997fast,voudouris1999guided}. However, GPGLS still inherits common limitations of Genetic Programming (GP), especially search stagnation and program bloat, where the population may lose effective search diversity while the evolved GP trees become increasingly large~\cite{luke2006comparison,silva2012operator,soule1998effects}.

Several mechanisms can address premature convergence and program growth in evolutionary search, including parsimony pressure, semantic diversity, archive-based search, restart strategies, and niching~\cite{li2016seeking,luke2006comparison,mahfoud1995niching,silva2012operator}. Niching is particularly relevant because it preserves multiple promising search regions instead of forcing the population toward a single dominant region. In evolutionary computation, fitness sharing, crowding, clearing, and archive-based diversity preservation have been widely used for this purpose~\cite{deb1989investigation,goldberg1987genetic,petrowski1996clearing}. Diversity management is also important in population-based VRP algorithms, where diverse routing structures or search patterns help reduce premature convergence~\cite{drake2020recent,vidal2022hgs}. These observations motivate the use of niching to preserve diverse GLS search heuristics in GPGLS.

However, directly applying conventional niching to GPGLS is non-trivial. Existing niching methods commonly define niches using fitness, program structure, or semantic outputs~\cite{mahfoud1995niching,petrowski1996clearing,vanneschi2014survey}. These criteria are not fully aligned with GPGLS because each GP individual represents a GLS utility function rather than a routing solution. In this setting, individuals with similar fitness may still guide GLS in different ways, while structurally different GP trees may induce similar search behaviour. Therefore, the relevant diversity in GPGLS should reflect not only objective values or tree syntax, but also the search behaviours produced during GLS execution.

To address this issue, this paper proposes GPGLS with Behaviour-based Niching (BN-GPGLS), a behaviour-based niching framework for GPGLS. BN-GPGLS defines niches using behaviour descriptors collected from GLS execution, reconstructs an archive of fitness-competitive, compact representatives in every generation, and uses these representatives under fixed or adaptive reproduction policies. The adaptive controller responds to training-fitness progress and standardised behavioural dispersion. BN-GPGLS aims to maintain useful search alternatives without modifying the routing fitness objective. It is a population-management intervention; final tree size is assessed as a separate outcome rather than sufficient evidence of control over the complete bloat process.
In summary, the main contributions of this paper are as follows:
\begin{itemize}
    \item Propose BN-GPGLS, a framework that equips GPGLS with behaviour-based niching by defining niches from GLS execution descriptors rather than fitness values or tree structure.

    \item Develop an adaptive archive-injection strategy that selects compact representatives from a behaviour archive and injects them as parent candidates when search stagnation is detected.

    \item Evaluate fixed and adaptive behaviour policies against no-archive and fitness-stratified controls, separating descriptive routing outcomes from paired statistical comparisons of final-population tree size.
\end{itemize}

\section{Background}
\subsection{Genetic Programming Guided Local Search for VRP}
\label{sec:gpgls}

VRPs have remained a central topic in combinatorial optimisation due to their applications in logistics, freight distribution, service planning, and supply-chain management~\cite{laporte2009fifty}. Exact methods can solve small VRP instances, but become computationally expensive on large-scale instances, where more scalable approaches are usually required~\cite{pessoa2020generic}. Heuristics, metaheuristics, and hyper-heuristics are therefore widely used for larger VRPs, as they can start from an initial feasible solution and rapidly improve its quality through problem-specific or automatically learned search rules~\cite{burke2013hyperheuristics,costa2023learning,costa2021evolutionary,drake2020recent,vidal2022hgs,wang2023explaining}.

GPGLS is an effective VRP-oriented hyper-heuristic framework that follows a two-level learning-to-optimise architecture for evolving GLS utility functions~\cite{liu2024gpgls}. A GLS-based evaluator solves training instances using a candidate utility function, while GP evolves a population of such functions according to evaluator feedback.

In GLS, the original objective is augmented by adaptive penalties on undesirable solution features, allowing local search to escape from local optima~\cite{gendreau2010handbook,liu2024gpgls}. In GPGLS, the initialisation procedure and local search operators are fixed, while the utility function is evolved. The evaluator therefore maps a GP individual and a set of training instances to a fitness value computed from the initial and final solution costs. Each individual is represented as a GP tree, and the population is updated through standard GP operations such as selection, crossover, and mutation~\cite{koza1994genetic}. After training, the best individual in the final generation is selected for final evaluation.

Despite its effectiveness, GPGLS still faces two limitations. First, the population may lose useful search diversity, leading to stagnation. Second, GP trees may grow without proportional performance improvement, causing program bloat~\cite{luke2006comparison,silva2012operator,soule1998effects}. Since GPGLS individuals are search heuristics rather than direct routing solutions, these issues should be analysed not only through fitness or tree structure, but also through the search behaviours induced during GLS.

\subsection{Niching and Diversity Preservation}
\label{sec:niching}
Niching is a diversity-preservation mechanism developed in evolutionary computation to maintain multiple promising regions within one population~\cite{goldberg1987genetic,li2016seeking,mahfoud1995niching}. Representative techniques include fitness sharing, crowding, clearing, and archive-based diversity preservation~\cite{deb1989investigation,goldberg1987genetic,li2016seeking,petrowski1996clearing}. Related diversity-oriented ideas have also been studied in semantic GP, novelty search, and quality-diversity optimisation~\cite{lehman2011abandoning,mouret2015illuminating,vanneschi2014survey}. In routing optimisation, population-diversity management is important in hybrid genetic search for VRPs, where local search is combined with diversity control to improve robustness~\cite{vidal2022hgs,vidal2013hybrid}.

In GPGLS, however, conventional niching criteria are not directly aligned with the role of an individual. Most niching methods measure similarity using fitness values, genotypes, solution structures, or program semantics~\cite{li2016seeking,mahfoud1995niching,vanneschi2014survey,petrowski1996clearing}, whereas a GPGLS individual represents a GLS utility function. Similar fitness values may therefore hide different GLS search behaviours, and different GP trees may induce similar behaviour. This motivates defining niches from behaviour descriptors collected during GLS execution.

\section{Proposed Method}
\subsection{Overall Framework and Fitness}
\label{sec:framework-overview}

\begin{figure}[!htbp]
  \centering
  \includegraphics[width=0.94\linewidth,height=0.69\textheight,keepaspectratio]{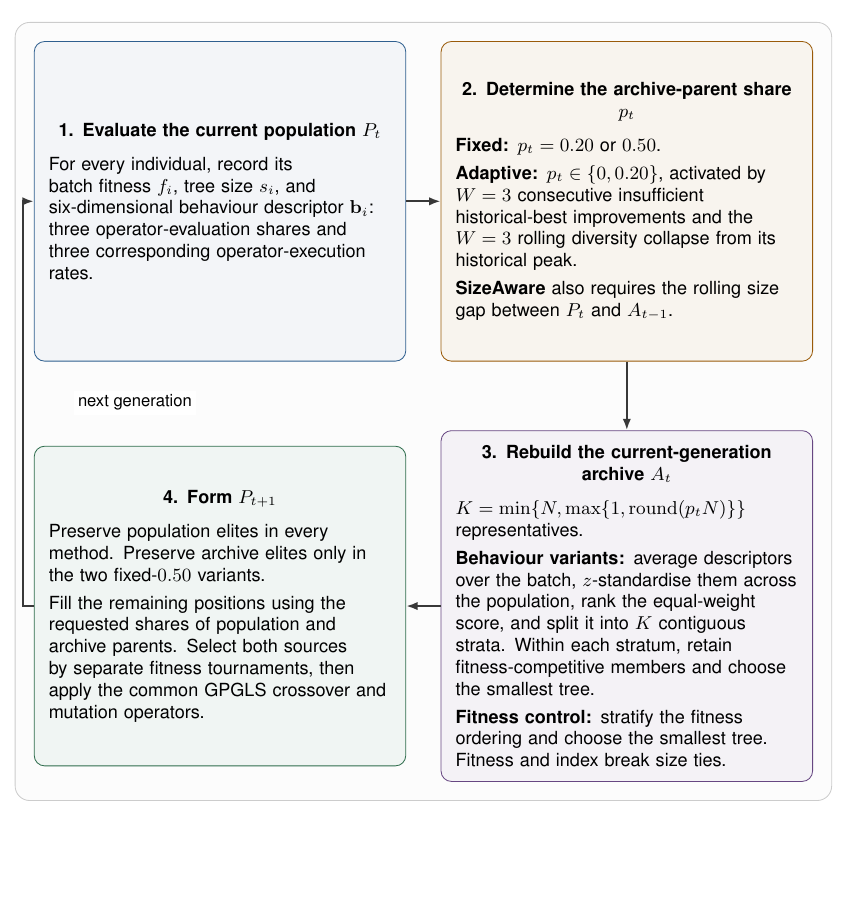}
  \caption{BN-GPGLS framework. Behaviour-based variants select competitive,
  compact representatives; the fitness-stratified control selects compact
  representatives without the median-fitness filter. Archive use follows the
  selected fixed or adaptive policy.}
  \label{fig:framework}
\end{figure}

Figure~\ref{fig:framework} shows BN-GPGLS. The GLS evaluator returns routing
fitness and behaviour descriptors for each GP tree. The archive is reconstructed
from the current evaluated population in every generation, rather than
accumulated across generations. Its representatives contribute to reproduction
according to the selected policy. The evaluator, routing objective, GLS
operators, and types of GP variation remain unchanged; parent allocation and,
for two fixed variants, the source of additional elites are modified.

Let $P_t=\{x_1,\ldots,x_N\}$ denote the generation-$t$ population, with tree
size $s_i$ for individual $x_i$. On a batch $B_t$ of $m$ training instances,
its minimised fitness is
\begin{equation}
  f_i=\frac{1}{m}\sum_{k\in B_t}
  \frac{C_{i,k}^{\mathrm{final}}-C_k^{\mathrm{init}}}{C_k^{\mathrm{init}}},
  \label{eq:fitness}
\end{equation}
where $C_k^{\mathrm{init}}$ is the common initial cost and
$C_{i,k}^{\mathrm{final}}$ is the cost obtained using $x_i$. More negative
fitness indicates greater relative improvement. Training-batch fitness drives
selection and the adaptive controller; monitoring-set results do not feed back
into training or individual selection.

\subsection{Behaviour Representation and Archive Construction}
\label{sec:behaviour-archive}

The descriptor observes three local-search operators: Lin--Kernighan,
cross-exchange, and relocation-chain. Let $e_{i,k,o}$ and $u_{i,k,o}$ denote
candidate moves evaluated and moves executed by operator $o\in\{1,2,3\}$
when $x_i$ is used on instance $k$. Evaluation shares and execution rates are
\begin{equation}
  a_{i,k,o}=\frac{e_{i,k,o}}{\sum_{j=1}^{3}e_{i,k,j}+10^{-9}},\qquad
  r_{i,k,o}=\frac{u_{i,k,o}}{e_{i,k,o}+10^{-9}}.
  \label{eq:operator-behaviour}
\end{equation}
The six-dimensional run descriptor is
\begin{equation}
  \mathbf b_{i,k}=(a_{i,k,1},a_{i,k,2},a_{i,k,3},r_{i,k,1},r_{i,k,2},r_{i,k,3}).
  \label{eq:run-descriptor}
\end{equation}
The evaluator returns these values rounded to six decimal places. Their
coordinate-wise batch mean is
$\bar{\mathbf b}_{i,t}=m^{-1}\sum_{k\in B_t}\mathbf b_{i,k}$.
No runtime feature or logarithmic count transformation is used.

For archive scoring, each coordinate is z-standardised across the current
population. If $\mu_{t,d}$ and $\sigma_{t,d}$ are its population mean and
standard deviation, respectively, then
\begin{equation}
  z_{i,t,d}=\frac{\bar b_{i,t,d}-\mu_{t,d}}{\sigma_{t,d}},\qquad
  \rho_{i,t}=\frac{1}{6}\sum_{d=1}^{6}z_{i,t,d},
  \label{eq:behaviour-score}
\end{equation}
where a standard deviation below $10^{-9}$ is replaced by one. Individuals
are sorted by the scalar score $\rho_{i,t}$ and divided into $K_t$ contiguous
strata of approximately equal size. For archive-parent share $p_t$,
\begin{equation}
  K_t=\min\{N,\max\{1,\operatorname{round}(p_tN)\}\}.
  \label{eq:stratum-count}
\end{equation}
Thus, $N=100$ yields 20 or 50 strata for the fixed shares, and one or 20 for
the inactive or active adaptive state. These niches are strata of a scalar
projection, not clusters in the full six-dimensional space.

Within behaviour stratum $\mathcal N_k$, the competitive set and its
representative are
\begin{align}
  \mathcal C_k&=\{x_i\in\mathcal N_k:
  f_i\leq\operatorname{median}_{x_j\in\mathcal N_k}f_j\},
  \label{eq:competitive-set}\\
  z_k&=\underset{x_i\in\mathcal C_k}{\arg\min}_{\mathrm{lex}}(s_i,f_i,i).
  \label{eq:representative}
\end{align}
The filter retains individuals no worse than their stratum median. Tree size
is then minimised, with fitness and population index breaking ties. The archive
is $A_t=\{z_1,\ldots,z_{K_t}\}$, replacing the preceding archive. Tree size is
not added to routing fitness. The fitness-niching control instead stratifies
directly by fitness and minimises $(s_i,f_i,i)$ within each stratum
\emph{without} the median-fitness filter.

\subsection{Adaptive Archive-Parent Allocation}
\label{sec:archive-injection}

Adaptive policies switch the archive-parent share between $p_{\mathrm{off}}=0$
and $p_{\mathrm{on}}=0.20$. The controller uses training signals with a fixed
window $W=3$; its thresholds do not adapt during a run.

\paragraph{Fitness signal.}
Let $F_t^\star$ be the stored historical-best reference before the scheduler
update and $f_t^{\min}$ the current minimum batch fitness. Relative improvement is
\begin{equation}
  \Delta_t^f=\max\left\{0,
  \frac{F_t^\star-f_t^{\min}}{\max\{|F_t^\star|,10^{-12}\}}\right\}.
  \label{eq:fitness-improvement}
\end{equation}
An improvement of at least $\varepsilon_f=10^{-4}$ updates the reference and
resets a consecutive-stagnation counter $c_t$; otherwise the counter increases
by one. At the first observation, the reference is set to $f_0^{\min}$ and
$c_0=0$. The fitness gate is $I_t^f=[c_t\geq W]$.

\paragraph{Behaviour signal.}
A separate preprocessing path robust-scales the six batch-mean coordinates
within the current population: each is centred on its median and divided by
its interquartile range, using the default scikit-learn
\texttt{RobustScaler} safeguard for zero or near-zero scales. Each scaled
coordinate is multiplied by $1/6$, giving $\breve{\mathbf b}_{i,t}$.
Average pairwise dispersion and its relative decline from the running peak are
\begin{align}
  D_t&=\frac{2}{N(N-1)}\sum_{i<j}
  \|\breve{\mathbf b}_{i,t}-\breve{\mathbf b}_{j,t}\|_2,\\
  h_t&=\operatorname{clip}\left(1-
  \frac{D_t}{\max\{\max_{0\leq\tau\leq t}D_\tau,10^{-9}\}},0,1\right).
  \label{eq:behaviour-collapse}
\end{align}
Let $\bar h_t$ be the mean of the latest up-to-$W$ values. The behaviour gate
is $I_t^b=[\bar h_t\geq0.12]$. This is a relative decline in a
population-standardised signal, not an absolute threshold on raw behavioural
spread.

\paragraph{Optional size signal.}
BN-SizeAware also compares mean tree size in the current population with that
in $A_{t-1}$, the archive available before the current reconstruction:
\begin{equation}
  g_t=\operatorname{clip}\left(
  \frac{\bar s(P_t)-\bar s(A_{t-1})}{\max\{\bar s(P_t),10^{-9}\}},0,1\right).
  \label{eq:size-gap}
\end{equation}
Set $g_t=0$ when no preceding archive exists. With $\bar g_t$ the mean of
the latest up-to-$W$ values, the size gate is $I_t^s=[\bar g_t\geq0.06]$.
BN-Adaptive activates $p_t=0.20$ when $I_t^f\land I_t^b$ holds;
BN-SizeAware additionally requires $I_t^s$. Otherwise $p_t=0$.
Changing training batches can affect the fitness signal, so these gates
operationalise intervention conditions rather than diagnose stagnation on a
fixed common instance set.

\paragraph{Reproduction.}
The share $p_t$ applies to non-elite parent slots. Every method copies five
population elites. FN-Fixed and BN-Fixed-0.5 additionally copy five archive
elites, then fill 90 slots with 45 population and 45 archive parents.
BN-Fixed-0.2 and active adaptive variants copy no archive elites and fill 95
slots with 76 population and 19 archive parents. Inactive adaptive variants
use 95 population parents. Each source uses a separate fitness tournament of
size three. The population pool excludes copied population elites; the
archive pool excludes copied archive elites when other representatives are
available. Selected parents are shuffled before crossover and mutation.
Algorithm~\ref{alg:niching} gives the adaptive procedure in execution order.

\begin{algorithm}[!htb]
\caption{Adaptive archive reconstruction and reproduction in BN-GPGLS}
\label{alg:niching}
\small
\begin{algorithmic}[1]
\Require Evaluated population $E_t=\{(x_i,f_i,\mathbf b_i,s_i)\}_{i=1}^{100}$,
preceding archive $A_{t-1}$, controller state
\Ensure Reconstructed archive $A_t$ and next population $P_{t+1}$
\State Update historical-best reference and stagnation counter $c_t$
\State Robust-scale batch-mean descriptors; compute $D_t$, $h_t$, and $\bar h_t$
\State Compute $g_t$ from $P_t$ and $A_{t-1}$, then its moving average $\bar g_t$
\State Set $p_t=0.20$ if the variant's gates hold; otherwise set $p_t=0$
\State Z-standardise batch-mean descriptors; compute scores $\rho_{i,t}$
\State Set $K_t$ by Eq.~\eqref{eq:stratum-count}; divide score ordering into $K_t$ strata
\For{each behaviour stratum $\mathcal N_k$}
  \State Retain individuals no worse than its median fitness
  \State Select $z_k$ by lexicographically minimising $(s_i,f_i,i)$
\EndFor
\State Reconstruct $A_t=\{z_1,\ldots,z_{K_t}\}$
\State Copy five population elites
\If{$p_t=0.20$}
  \State Select 76 population and 19 archive parents by separate fitness tournaments
\Else
  \State Select 95 population parents by fitness tournaments
\EndIf
\State Shuffle selected parents and apply crossover and mutation
\State \Return $A_t,P_{t+1}$
\end{algorithmic}
\end{algorithm}

\FloatBarrier
\section{Experimental Design}
\label{sec:exp-settings}

\subsection{Data and Training Protocol}

The experiments use generated capacitated VRP instances with 200 customers
and Euclidean distances. The XML-style generator follows the design factors
used in XML100 and the Uchoa benchmark framework~\cite{queiroga202110,uchoa2017new}.
Each 90-instance set contains one instance for every combination of three
customer-position distributions, six demand distributions, and five average
route-size categories, as specified in Table~\ref{tab:data}. The depot-position
factor is random. Independent realisations of the same factor combinations
form the training and monitoring sets.

\begin{table}[!htb]
\centering
\small
\caption{Factor levels in each 90-instance dataset.}
\label{tab:data}
\begin{tabular}{@{}>{\raggedright\arraybackslash}p{0.24\textwidth}>{\raggedright\arraybackslash}p{0.71\textwidth}@{}}
\toprule
Factor & Levels \\
\midrule
Customer position & Random; clustered; half random and half clustered \\
Demand distribution & $1$--$10$; $5$--$10$; $1$--$100$; $50$--$100$;
quadrant-dependent ($1$--$50$ or $51$--$100$); many small values
($1$--$10$) and few large values ($50$--$100$) \\
Target customers per route & Short ($5$--$8$); medium ($8$--$12$);
long ($12$--$16$); very long ($16$--$25$); ultra long ($25$--$50$) \\
\bottomrule
\end{tabular}
\end{table}

Integer coordinates are sampled from a $1000\times1000$ plane, with two to
six cluster seeds for clustered layouts. Vehicle capacity is the larger of
the maximum customer demand and the sampled target route size multiplied by
mean demand, rounded upward. The 90 training instances are shuffled once with
schedule seed 2026 and divided into 18 non-repeated batches of five. All runs
and methods use this same batch sequence.

Each method is run with 30 random seeds matched across methods. The
best training-fitness individual from each generation is saved and evaluated
on all 90 monitoring instances \emph{after} training. This yields 18 monitoring
blocks per run; the last block uses the final generation's best individual
and supplies the final routing results. Monitoring does not select individuals
or alter the controller. The separate set therefore measures same-scale,
same-generator-family generalisation, with its repeated generation-wise
evaluations used as a monitoring diagnostic.

\subsection{Compared Methods and Settings}

Table~\ref{tab:methods} defines the six compared policies. The GPGLS control
uses the same training configuration as the niching variants, not the original
GPGLS paper's experimental schedule. FN-Fixed and BN-Fixed-0.5 share archive
allocation and elite sources but differ in both stratification and the
median-fitness filter. BN-Fixed-0.2 and BN-Adaptive contrast continuous with
triggered use at the same active share and elite source. BN-SizeAware adds
only the size gate to BN-Adaptive.

\begin{table}[!htb]
\centering
\small
\caption{Compared methods. All policies retain five population elites.}
\label{tab:methods}
\setlength{\tabcolsep}{3pt}
\begin{tabular}{@{}>{\raggedright\arraybackslash}p{0.20\textwidth}>{\raggedright\arraybackslash}p{0.32\textwidth}>{\raggedright\arraybackslash}p{0.23\textwidth}>{\raggedright\arraybackslash}p{0.15\textwidth}@{}}
\toprule
Method & Strata and representative & Archive-parent share & Archive elites \\
\midrule
GPGLS control & No archive & $0$ & $0$ \\
FN-Fixed & Fitness; compact, no median filter & Fixed $0.50$ & $5$ \\
BN-Fixed-0.2 & Behaviour; competitive-compact & Fixed $0.20$ & $0$ \\
BN-Fixed-0.5 & Behaviour; competitive-compact & Fixed $0.50$ & $5$ \\
BN-Adaptive & Behaviour; competitive-compact & Adaptive $0/0.20$ & $0$ \\
BN-SizeAware & Behaviour; competitive-compact & Adaptive $0/0.20$ with size gate & $0$ \\
\bottomrule
\end{tabular}
\end{table}

Table~\ref{tab:hyperparams} gives the shared training settings and BN
parameters. The listed ratios, window, and thresholds remain fixed throughout
each run; adaptation changes only the active archive-parent share. Monitoring
results are not inputs to this process. This study evaluates the listed
configuration rather than a parameter-sensitivity analysis.

\begin{table}[!htb]
\centering
\small
\caption{GP training and behaviour-niching settings.}
\label{tab:hyperparams}
\begin{tabular}{@{}ll@{}}
\toprule
Parameter & Setting \\
\midrule
Population size; tournament size & $100$; $3$ \\
Crossover; mutation probability & $0.80$; $0.15$ \\
Population elite ratio & $0.05$ \\
Initialisation; initial depth & Ramped half-and-half; $2$--$6$ \\
Generations; seed-matched runs & $18$; $30$ \\
Training / monitoring instances & $90$ / $90$ \\
Training batch size & $5$ per generation \\
Evaluator limits & 5,000 iterations; 20-s CPU-time limit \\
Behaviour descriptor & 3 evaluation shares; 3 execution rates \\
Fixed archive-parent shares & $0.20$, $0.50$ \\
Adaptive inactive / active shares & $0$ / $0.20$ \\
Controller window $W$ & $3$ \\
Fitness threshold $\varepsilon_f$ & $10^{-4}$ \\
Relative-dispersion threshold $\tau_b$ & $0.12$ \\
Size-gap threshold $\tau_s$ & $0.06$ \\
\bottomrule
\end{tabular}
\end{table}

\subsection{Metrics and Statistical Analysis}

Final routing cost is first averaged over 30 runs for each method--instance
pair. The six method means are ranked within each monitoring instance, with
average ranks for ties, and averaged over the 90 instances. Overall mean and
median costs summarise the 90 instance-level means. These routing outcomes
are descriptive; no run-level inferential routing-cost comparison is made.

Each generation-wise monitoring point averages 30 generation-best trees on
the common 90 instances. Least-squares trends through the 18 points provide
visual summaries, not slope tests or tests of recovery following activation.

For tree size, each run contributes the median node count across its 100
final-population trees. Across-run medians and interquartile ranges summarise
these 30 outcomes. Five two-sided paired Wilcoxon signed-rank comparisons
against the GPGLS control use the matched random seeds. Holm adjustment is
applied across the five comparisons using unrounded raw $p$-values, with a
$p$-value threshold of 0.05 on the adjusted values. The tests address
final-population size, not routing cost or the complete growth trajectory.

\FloatBarrier
\section{Results and Discussion}
\subsection{Final Routing Cost and Average Rank}

Table~\ref{tab:test-rank} reports results for the last-generation individuals
on the monitoring set. BN-Adaptive has the best average rank, whereas
BN-SizeAware has the lowest mean and median costs. The GPGLS control has the
worst rank and mean, but a lower median than the three fixed archive policies.

\begin{table}[!htb]
\centering
\small
\caption{Last-generation routing cost and average rank on the 90 monitoring
instances. Lower values are better.}
\label{tab:test-rank}
\begin{tabular}{lrrr}
\toprule
Method & Average rank & Mean cost & Median cost \\
\midrule
BN-Adaptive & \textbf{2.79} & 23,870.18 & 21,412.57 \\
BN-SizeAware & 3.01 & \textbf{23,869.30} & \textbf{21,410.15} \\
BN-Fixed-0.5 & 3.48 & 23,869.32 & 21,424.62 \\
BN-Fixed-0.2 & 3.72 & 23,873.38 & 21,420.07 \\
FN-Fixed & 3.83 & 23,873.59 & 21,420.03 \\
GPGLS control & 4.18 & 23,879.60 & 21,416.42 \\
\bottomrule
\end{tabular}
\end{table}

The cost differences are small. The largest reduction from the control's
mean of 23,879.60 is 10.30 units for BN-SizeAware, or 0.043\%. Its median is
6.27 units (0.029\%) lower. The rank ordering therefore describes the
observed comparisons, not statistically established or necessarily practically
important routing-cost superiority.

Under matched fixed-0.50 allocation and elitism, BN-Fixed-0.5 has a better
rank and mean than FN-Fixed. BN-Adaptive has a better rank, mean, and median
than BN-Fixed-0.2, which shares its elite source and active parent share.
Adding the size gate changes this trade-off: BN-SizeAware improves the raw
costs slightly but not the average rank. These are contrasts among the tested
policies, not isolated causal effects of every component.

\subsection{Final-Population GP Tree Size}

\begin{figure}[!htb]
  \centering
  \includegraphics[width=0.94\textwidth]{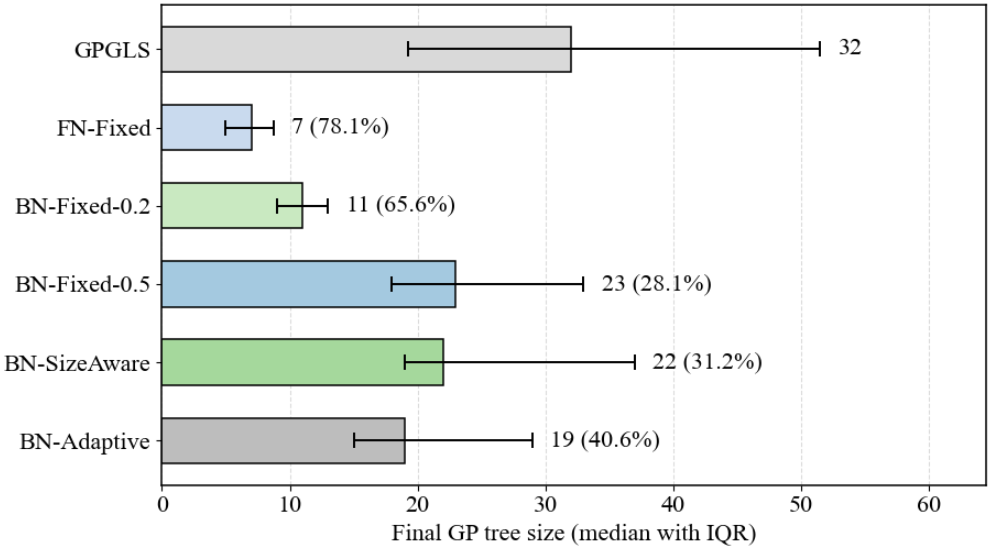}
  \caption{Final-population tree size. Each run contributes the median node
  count of its 100 final-population trees; bars show the across-run median,
  error bars span the interquartile range, and labels give the median
  reduction from the GPGLS control.}
  \label{fig:size}
\end{figure}

Table~\ref{tab:size} gives final-population tree-size summaries and paired
test results. Figure~\ref{fig:size} visualises the same 30-run summaries.

\begin{table}[!htb]
\centering
\footnotesize
\setlength{\tabcolsep}{3pt}
\caption{Final-population median tree size across 30 runs and paired Wilcoxon
comparisons against the GPGLS control. Holm adjustment covers all five comparisons.}
\label{tab:size}
\begin{tabular}{lrrrr}
\toprule
Method & Median [Q1, Q3] & Reduction & Raw $p$ & Holm $p$ \\
\midrule
GPGLS control & $32$ [$19.25$, $51.50$] & -- & -- & -- \\
FN-Fixed & $7$ [$5.00$, $8.75$] & $78.13\%$ & $2.55\times10^{-6}$ & $1.02\times10^{-5}$ \\
BN-Fixed-0.2 & $11$ [$9.00$, $13.00$] & $65.63\%$ & $1.91\times10^{-6}$ & $9.57\times10^{-6}$ \\
BN-Adaptive & $19$ [$15.00$, $29.00$] & $40.63\%$ & $2.27\times10^{-3}$ & $6.81\times10^{-3}$ \\
BN-SizeAware & $22$ [$19.00$, $37.00$] & $31.25\%$ & $1.79\times10^{-2}$ & $3.57\times10^{-2}$ \\
BN-Fixed-0.5 & $23$ [$18.25$, $32.00$] & $28.13\%$ & $2.52\times10^{-2}$ & $3.57\times10^{-2}$ \\
\bottomrule
\end{tabular}
\end{table}

All five archive policies have lower across-run medians than the GPGLS
control, and all five paired comparisons remain below 0.05 after Holm
adjustment. FN-Fixed yields the smallest descriptive median, seven nodes,
against 32 for the control. BN-Adaptive has a median of 19 nodes, a 40.63\%
reduction. The tests compare each policy with the control, not the niching
policies directly with one another.

The smallest trees do not coincide with the best routing rank. FN-Fixed is
most compact but has a weaker average rank than all four behaviour policies.
BN-Adaptive gives the best descriptive rank with a lower size median than
the control; BN-SizeAware gives slightly lower raw costs and a 22-node median.
These results describe different program-size and solution-quality trade-offs.
Because compact selection is shared by all archive policies and some policies
also change elitism, the comparisons do not isolate a behaviour-specific
cause of the size reductions.

\subsection{Generation-wise Monitoring}

\begin{figure}[!htb]
  \centering
  \includegraphics[width=\textwidth]{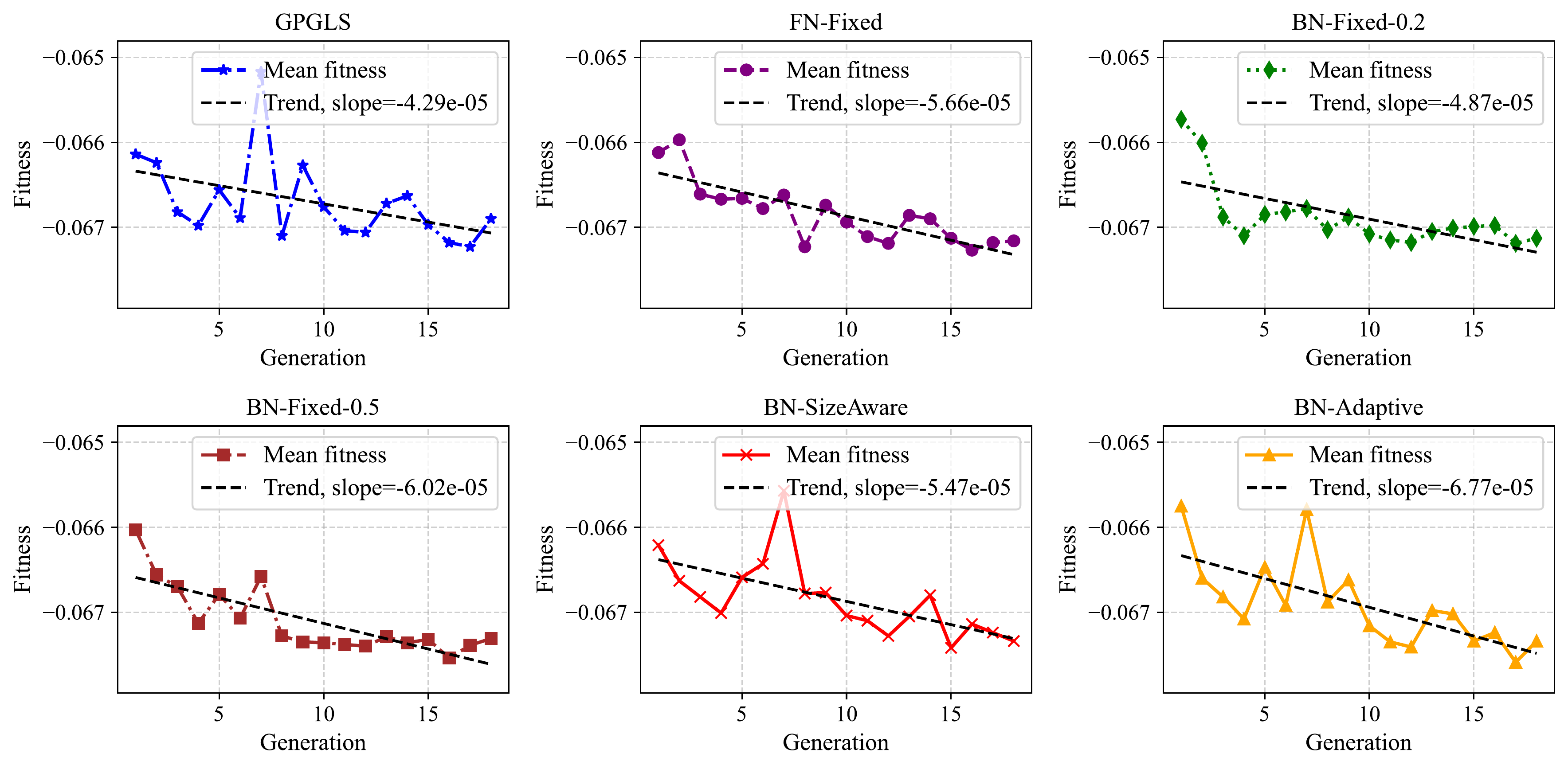}
  \caption{Generation-wise monitoring-set fitness. Each point averages the
  generation-best trees over 30 runs and 90 monitoring instances. Fitted
  least-squares trends are descriptive summaries, not statistical tests.}
  \label{fig:validation}
\end{figure}

Figure~\ref{fig:validation} shows monitoring fitness for the best
training-fitness tree of each generation. Each mean covers 2,700 evaluations
(30 runs on 90 instances). Lower values indicate greater relative improvement
under Eq.~\eqref{eq:fitness}.

The fitted slopes are $-6.77\times10^{-5}$ for BN-Adaptive,
$-6.02\times10^{-5}$ for BN-Fixed-0.5, $-5.66\times10^{-5}$ for FN-Fixed,
$-5.47\times10^{-5}$ for BN-SizeAware, $-4.87\times10^{-5}$ for
BN-Fixed-0.2, and $-4.29\times10^{-5}$ for the GPGLS control. BN-Adaptive
has the strongest whole-series decline, and BN-Fixed-0.5 the strongest decline
among fixed archive policies. FN-Fixed declines more strongly than
BN-Fixed-0.2 and BN-SizeAware, so the curves do not support a uniform trend
advantage for behaviour-based variants. Across runs, BN-Adaptive activates
archive parents for a mean of 5.10 of the 17 breeding transitions and
BN-SizeAware for 4.63. These counts establish selective use, but the
whole-series trends do not attribute recovery after activation to archive use.

\FloatBarrier
\subsection{Scope and Limitations}

The descriptor excludes timing, temporal order, and operator transitions, and
scalar projection can merge distinct behaviours. Refitting the robust scaler
each generation makes dispersion approximately invariant to coordinate-wise
location and scale changes, so the controller does not directly measure
absolute raw-behaviour contraction.

The evaluation covers one parameter configuration, one batch sequence, and
one family of 200-customer instances. Transfer to other scales or distributions
remains untested. Routing outcomes are descriptive; tree-size tests compare
policies only against the control and do not isolate every component or
measure complete growth trajectories. Cross-generation archive retention is
also outside the design.

Once descriptors have been collected, batch aggregation and scaling require
$O(Nq)$ work, sorting requires $O(N\log N)$, and adaptive pairwise dispersion
requires $O(N^2q)$, with $q=6$. These costs are additional to GLS evaluation.
No controlled timing decomposition was performed, so they do not quantify
realised overhead or customer-scale performance.

\section{Conclusion}

BN-GPGLS extends learned GLS utility functions with behaviour-based population
management. Six operator descriptors define strata from which competitive,
compact representatives reconstruct an archive. Fixed and adaptive policies
then allocate archive parents without changing routing fitness.

BN-Adaptive obtains the best descriptive monitoring-set rank and BN-SizeAware
the lowest raw costs, although cost differences are small. All five archive
policies yield lower final-population size medians than the control, with
significant Holm-adjusted paired comparisons. These findings support useful
trade-offs in the evaluated setting, not universal routing superiority or a
behaviour-only explanation for smaller trees.

Future work should investigate temporal descriptors and operator transitions,
parameter sensitivity, controlled component comparisons, activation-conditioned
trajectories, and larger or differently distributed routing instances.
Cross-generation archives and direct computational-overhead measurements
would further clarify when these policies are useful.

\bibliographystyle{splncs04}
\bibliography{ref}
\end{document}